\documentclass[10pt,twocolumn,letterpaper]{article}

\usepackage{wacv}              %

\usepackage{booktabs}
\usepackage{caption}
\usepackage{pifont}  
\usepackage{amssymb}
\usepackage{graphicx}
\usepackage{amsmath}
\usepackage{booktabs}
\usepackage{mathtools}
\usepackage{xcolor}
\usepackage{notes2bib}
\graphicspath{ {./images/} }
\usepackage{multirow}
\usepackage{rotating}
\usepackage{tabularx}
\usepackage{arydshln}
\usepackage{wrapfig}
\usepackage{caption}
\usepackage{enumitem}
\usepackage{adjustbox}
\usepackage{dblfloatfix}
\newcommand*\samethanks[1][\value{footnote}]{\footnotemark[#1]}
\usepackage{siunitx}

\usepackage[pagebackref,breaklinks,colorlinks]{hyperref}

\usepackage[capitalize]{cleveref}
\crefname{section}{Sec.}{Secs.}
\Crefname{section}{Section}{Sections}
\Crefname{table}{Table}{Tables}
\crefname{table}{Tab.}{Tabs.}
\Crefname{enumi}{Step}{Steps}
\crefname{enumi}{Step}{Steps}

\def\wacvPaperID{458} %
\def\confName{WACV}
\def\confYear{2024}

\begin{document}

\title{GEFF: Improving Any Clothes-Changing Person ReID Model using Gallery Enrichment with Face Features}
\author{Daniel Arkushin\thanks{These authors contributed equally}$^{*1}$ \qquad Bar Cohen\samethanks$^{*2}$ \qquad Shmuel Peleg$^1$ \qquad Ohad Fried$^2$ \\
$^1$The Hebrew University of Jerusalem $^2$Reichman University\\ 
\small\url{http://www.vision.huji.ac.il/reface/} \\
{\tt\small \{daniel.arkushin, peleg\}@mail.huji.ac.il } {\tt\small bar.cohen01@post.runi.ac.il } {\tt\small ofried@runi.ac.il} \\
}

\maketitle

\def\ShowNotes{}

\newcommand{\ignorethis}[1]{}
\newcommand{\redund}[1]{#1}

\newcommand{\apriori    }     {\textit{a~priori}}
\newcommand{\aposteriori}     {\textit{a~posteriori}}
\newcommand{\perse      }     {\textit{per~se}}
\newcommand{\naive      }     {{na\"{\i}ve}}

\newcommand{\Identity   }     {\mat{I}}
\newcommand{\Zero       }     {\mathbf{0}}
\newcommand{\Reals      }     {{\textrm{I\kern-0.18em R}}}
\newcommand{\isdefined  }     {\mbox{\hspace{0.5ex}:=\hspace{0.5ex}}}
\newcommand{\texthalf   }     {\ensuremath{\textstyle\frac{1}{2}}}
\newcommand{\half       }     {\ensuremath{\frac{1}{2}}}
\newcommand{\third      }     {\ensuremath{\frac{1}{3}}}
\newcommand{\fourth     }     {\ensuremath{\frac{1}{4}}}

\newcommand{\Lone} {\ensuremath{L_1}}
\newcommand{\Ltwo} {\ensuremath{L_2}}

\newcommand{\mat        } [1] {{\text{\boldmath $\mathbit{#1}$}}}
\newcommand{\Approx     } [1] {\widetilde{#1}}
\newcommand{\change     } [1] {\mbox{{\footnotesize $\Delta$} \kern-3pt}#1}

\newcommand{\Order      } [1] {O(#1)}
\newcommand{\set        } [1] {{\lbrace #1 \rbrace}}
\newcommand{\floor      } [1] {{\lfloor #1 \rfloor}}
\newcommand{\ceil       } [1] {{\lceil  #1 \rceil }}
\newcommand{\inverse    } [1] {{#1}^{-1}}
\newcommand{\transpose  } [1] {{#1}^\mathrm{T}}
\newcommand{\invtransp  } [1] {{#1}^{-\mathrm{T}}}
\newcommand{\relu       } [1] {{\lbrack #1 \rbrack_+}}

\newcommand{\abs        } [1] {{| #1 |}}
\newcommand{\Abs        } [1] {{\left| #1 \right|}}
\newcommand{\norm       } [1] {{\| #1 \|}}
\newcommand{\Norm       } [1] {{\left\| #1 \right\|}}
\newcommand{\pnorm      } [2] {\norm{#1}_{#2}}
\newcommand{\Pnorm      } [2] {\Norm{#1}_{#2}}
\newcommand{\inner      } [2] {{\langle {#1} \, | \, {#2} \rangle}}
\newcommand{\Inner      } [2] {{\left\langle \begin{array}{@{}c|c@{}}
                               \displaystyle {#1} & \displaystyle {#2}
                               \end{array} \right\rangle}}

\newcommand{\twopartdef}[4]
{
  \left\{
  \begin{array}{ll}
    #1 & \mbox{if } #2 \\
    #3 & \mbox{if } #4
  \end{array}
  \right.
}

\newcommand{\fourpartdef}[8]
{
  \left\{
  \begin{array}{ll}
    #1 & \mbox{if } #2 \\
    #3 & \mbox{if } #4 \\
    #5 & \mbox{if } #6 \\
    #7 & \mbox{if } #8
  \end{array}
  \right.
}

\newcommand{\len}[1]{\text{len}(#1)}

\newlength{\w}
\newlength{\h}
\newlength{\x}

\definecolor{darkred}{rgb}{0.7,0.1,0.1}
\definecolor{darkgreen}{rgb}{0.1,0.6,0.1}
\definecolor{cyan}{rgb}{0.7,0.0,0.7}
\definecolor{otherblue}{rgb}{0.1,0.4,0.8}
\definecolor{maroon}{rgb}{0.76,.13,.28}
\definecolor{burntorange}{rgb}{0.81,.33,0}

\ifdefined\ShowNotes
  \newcommand{\colornote}[3]{{\color{#1}\textbf{#2} #3\normalfont}}
\else
  \newcommand{\colornote}[3]{}
\fi

\newcommand {\todo}[1]{\colornote{cyan}{TODO}{#1}}
\newcommand {\ohad}[1]{\colornote{otherblue}{OF:}{#1}}
\newcommand {\shmuel}[1]{\colornote{darkgreen}{SP:}{#1}}
\newcommand {\barc}[1]{\colornote{burntorange}{BC:}{#1}}
\newcommand {\daniel}[1]{\colornote{maroon}{DA:}{#1}}
\newcommand {\rotem}[1]{\colornote{red}{RS:}{#1}}

\newcommand {\reqs}[1]{\colornote{red}{\tiny #1}}

\newcommand {\new}[1]{\colornote{red}{#1}}

\newcommand*\rot[1]{\rotatebox{90}{#1}}

\newcommand {\newstuff}[1]{#1}

\newcommand\todosilent[1]{}
\newcommand{\Gface}{$G_{face}$}
\newcommand{\Genriched}{$G_{enriched}$}
\newcommand{\Genrichedi}{$G_{enriched_i}$}

\newcommand{\woBGmask}{{w/o~bg~\&~mask}}
\newcommand{\woMask}{{w/o~mask}}

\providecommand{\keywords}[1]
{
  \textbf{\textit{Keywords---}} #1
}

\newcommand{\GAN}{\textit{GAN}}
\newcommand{\data}{\mathit{data}}
\newcommand{\unionGAN}{\textsc{UnionGAN}\xspace}
\newcommand {\ganArrow}[2]{\ensuremath{\GAN_{{#1} \rightarrow {#2}}}}
\newcommand {\gan}[1]{\ensuremath{\GAN_{#1}}}
\newcommand{\DALLE}{{DALL$\cdotBB$E}}

\begin{abstract}
In the Clothes-Changing Re-Identification (CC-ReID) problem, given a query sample of a person, the goal is to determine the correct identity based on a labeled gallery in which the person appears in different clothes. Several models tackle this challenge by extracting clothes-independent features. However, the performance of these models is still lower for the clothes-changing setting compared to the same-clothes setting in which the person appears with the same clothes in the labeled gallery. As clothing-related features are often dominant features in the data, we propose a new process we call Gallery Enrichment, to utilize these features. 
In this process, we enrich the original gallery by adding to it query samples based on their face features, using an unsupervised algorithm. 
Additionally, we show that combining ReID and face feature extraction modules alongside an enriched gallery results in a more accurate ReID model, even for query samples with new outfits that do not include faces. 
Moreover, we claim that existing CC-ReID benchmarks do not fully represent real-world scenarios, and propose a new video CC-ReID dataset called \textbf{\textit{42Street}}, based on a theater play that includes crowded scenes and numerous clothes changes.
When applied to multiple ReID models, our method (GEFF) achieves an average improvement of 33.5\% and 6.7\% in the Top-1 clothes-changing metric on the PRCC and LTCC benchmarks. Combined with the latest ReID models, our method achieves new SOTA results on the PRCC, LTCC, CCVID, LaST and VC-Clothes benchmarks and the proposed \textbf{\textit{42Street}} dataset.

\end{abstract}

\section{Introduction}
\label{sec:intro}

Person re-identification (ReID) aims to match the same people appearing at different times and locations.
Given samples of people of interest --- commonly referred to as a \textit{gallery}, and unlabeled query samples, the goal is to predict the correct label (\ie person ID) for every query sample based on the given gallery. Existing ReID models tend to perform poorly when re-identifying the same people over a prolonged time due to appearance changes such as different clothes and hairstyles\cite{reid_aim}. Moreover, the performance of models that try to extract clothes-independent features such as body shape \cite{silhouettes}, contours \cite{prcc,m2net}, or gait \cite{gait-FVG,gait-survey}, is subpar compared to same-clothes settings,
 as clothes are often the most dominant features \cite{CAL,reid_aim}. To address the clothes-changing problem we introduce a simple process which we refer to as \textit{Gallery Enrichment}. In this process, we use the gallery data to automatically add to it parts of the query data where people appear in different outfits. Extending the gallery in this manner results in an enriched gallery that increases the chances of finding a correct match. This is done by an unsupervised algorithm that uses the similarity between the faces in the query and gallery samples.

\begin{figure*}
    \centering
    \includegraphics[width=\linewidth]{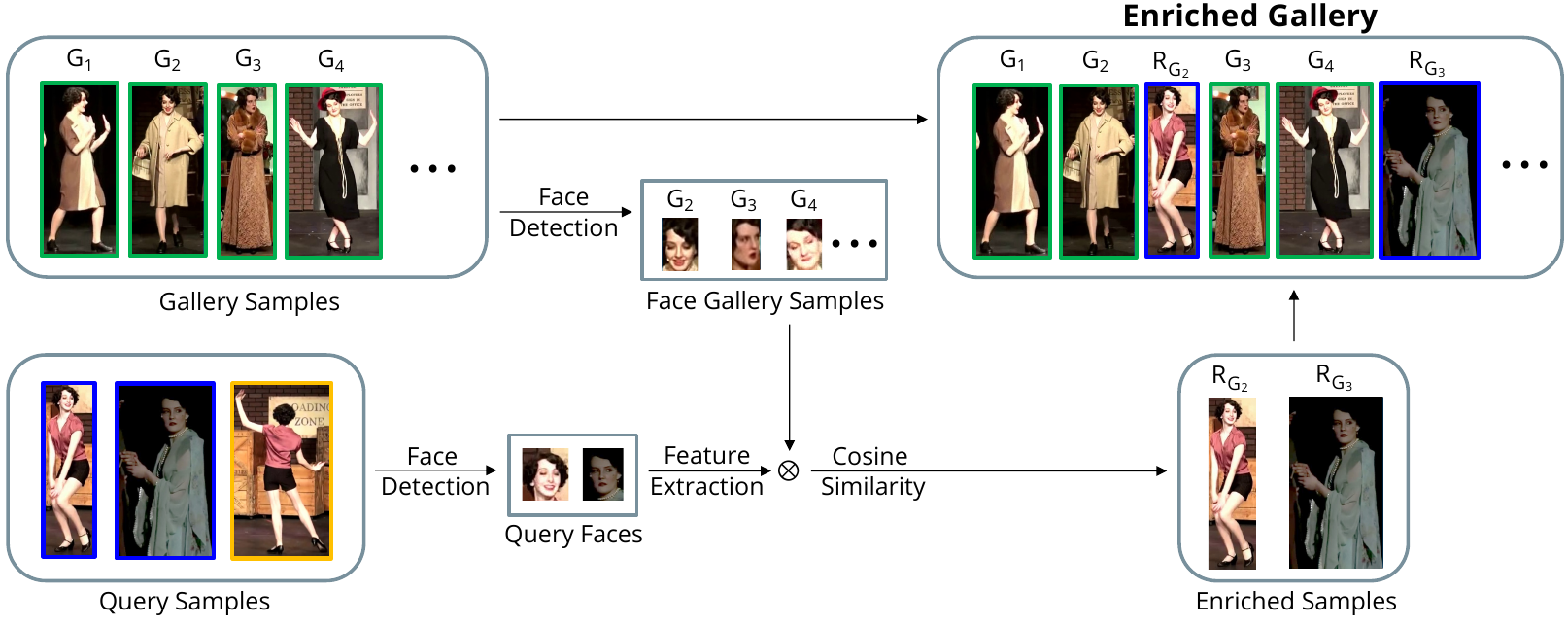}
    \caption{\textbf{The Gallery Enrichment Process.} For every gallery sample in which a face was detected, a face feature vector is extracted to create a face gallery. Then, for every query sample in which a face was detected, a reference to the closest sample in the gallery is saved, based on face similarity. Finally, an enriched gallery is created by extending the original gallery with the enriched samples. Colored frames were added for visualization purposes; green indicates a gallery sample and blue a query sample that is added to the enriched gallery. Notice that query samples that do not include faces (yellow frame) will not be added to the gallery. However, as an outcome of the enrichment process, during the ReID inference (not illustrated) it will be more likely to find a correct match for such queries.
    }
    \label{fig:gallery-enrichment}
\end{figure*}

 As face features provide an accurate prediction for the query samples that include faces, this algorithm results in an enriched gallery with minimal errors (\cref{fig:gallery-enrichment}). Once enriched, the gallery allows the ReID model, which relies on appearance-related features, to correctly predict the identity of a person with previously unseen outfits, even if the query sample does not include a face.
In addition to using faces in the gallery enrichment process, we claim that integrating a face feature extraction module during inference is beneficial for the results of ReID, and introduce a new method that combines pre-trained face features extraction and ReID modules alongside an enriched gallery. We call this method GEFF --- Gallery Enrichment with Face Features.

We claim that current video CC-ReID benchmarks do not include enough cases of occlusions, various illumination conditions, and multiple clothes and hairstyle changes. Therefore, we introduce the  \textbf{\textit{42Street}} dataset, curated from a theater play, as many theater plays include these challenges. 

Extensive experiments show that GEFF improves the performance of the evaluated ReID models, on 5 CC-ReID benchmarks and the \textbf{\textit{42Street}} dataset.

\section{Related Work}
\subsection{Person Re-Identification} \label{reid_background}
The common inference process of person ReID can be seen as an instance retrieval problem. Given gallery and query samples, the goal is to classify each query sample correctly. First, feature vectors for all gallery samples are extracted by applying a feature-extraction model. Next, given an unseen query sample, the distances between its feature vector and the gallery feature vectors are computed. Finally, the predicted label of the query sample is defined as the label of the gallery feature vector that is closest to the query.

\textbf{Image-based and Video-based ReID}
In image-based ReID datasets, every data sample is a single-person crop. Over the years, multiple image-based models have been developed \cite{CTL, fast-reid, BagTricks, PyrNet}, and they achieve impressive results on same-clothes image-based benchmarks. In video-based ReID datasets, every data sample is a sequence of crops, \ie a track, and the video ReID model produces a single feature vector to represent the entire sequence by using the spatio-temporal information in the sequence \cite{gait-survey,CAL,VideoReID,denseIL}.

\textbf{Clothes-changing ReID} 
In this setting, a person in the query set might wear different clothes from the gallery set. Some models try to extract clothes-independent features by modeling body shape and gait using skeletons \cite{ltcc}, silhouettes \cite{silhouettes}, and contour sketches \cite{prcc}. M2Net \cite{m2net} uses contour images and human-parsing images to extract meaningful features.  CAL \cite{CAL}, proposes a Clothes-based Adversarial Loss to mine clothes-independent features, and uses the video input to extract spatio-temporal patterns. AIM \cite{reid_aim} utilizes a causality-based auto-intervention model to mitigate clothing bias and CCFA \cite{CCFA} implicitly augments clothes-changing data in the feature space. 
A concurrent work, IGCL \cite{IGCL}, applies vision transformers to provide direct supervision to learn identity-specific features. Since the accuracy of these models is lower under the clothes-changing setting compared to the same-clothes setting, we suggest a process that partially converts the clothes-changing setting into the same-clothes setting, by building an enriched gallery and using face features during inference.

\subsection{Face Feature Extraction}
Face feature extraction is the process of detecting and identifying specific facial features from images or videos. Early works in the field such as Viola-Jones \cite{viola-Jones} and Local Binary Pattern \cite{local_binary_pattern} laid the foundations for more recent methods such as Face Attention Network\cite{wang2017face} and the Insightface\cite{deng2018arcface,deng2018menpo,Deng2020CVPR,deng2020subcenter} library, which use deep learning to extract facial features and surpass human performance\cite{face_recognition_survery}.

\textbf{Using Face Features for ReID}
Several studies attempted utilizing face features for the task of person ReID, using various deep learning techniques \cite{Face-reID-thermal,face_similarty_pei_li, Deep_Facial_Person,towards_facial_ReID}. While these works try to predict an identity based on face features only, as we show in our work, face features are insufficient on their own since not all query samples contain faces. Therefore, we propose a model that combines both face and ReID modules. Another work, 3APF \cite{VC-Clothes}, combines a holistic feature extractor (ReID part) and a local face feature extractor (face part) on the feature vectors space. In our method, we propose to create a score vector for each model separately and then combine them into a final score vector. Additionally, we use face features to build the enriched gallery. In our experiments, we show that we outperform 3APF on VC-Clothes\cite{VC-Clothes}, the dataset they publish.

\section{Method}
In order to address the clothes-changing ReID problem, our method enriches the gallery using an unsupervised process (\cref{gallery-creation}) and combines a pre-trained ReID module together with a pre-trained face feature extraction module (\cref{subsec:combined-reid-and-face}). Both elements work in conjunction and define our method which can be applied to any ReID model and works with image-based and video-based settings.

\subsection{Creating an Enriched Gallery} \label{gallery-creation} 
The objective of most ReID models is to produce a feature extraction function that generates a feature vector for each data sample. Given two different data samples of the same person, the feature extraction function is expected to generate feature vectors with a lower distance compared to two samples of different persons. The richer the gallery is with samples that are similar to the query set, the more likely it is to find a correct gallery match.
Hence, we propose the following unsupervised algorithm to enrich a given gallery from the query using face features, when available (\cref{fig:gallery-enrichment}).
Given gallery and query samples:
\begin{enumerate}[leftmargin=*]

\item \label{item:labeled} We first apply a face detector on all gallery samples. We then build a face gallery by applying a face feature extractor on every sample in which a face was detected.
\item For every query sample in which a face was detected,
we save a reference to the most similar gallery sample by computing its face feature vector and comparing it to the face gallery from \Cref{item:labeled} using cosine-similarity. \label{item:unlabeled}

\item We create an enriched gallery by extending the original gallery with the queries from \Cref{item:unlabeled}.
\end{enumerate}

During evaluation, the references to the original gallery samples are used to determine the predicted identity of a given query sample.

\subsubsection{Matching Face to Pose Estimation} \label{face-to-pose} 

\begin{figure}

    \centering
    \setlength{\tabcolsep}{5pt}
    \begin{tabular}{ccc}
        \frame{\includegraphics[trim=3 3 3 3,clip,width=20mm]{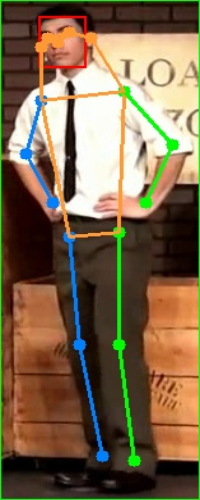}} &
        \frame{\includegraphics[trim=3 3 3 3,clip,width=20mm]{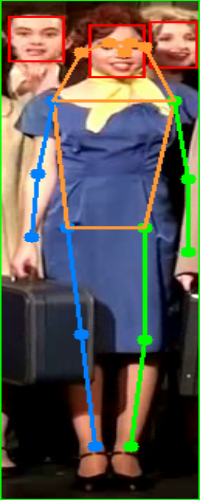}} &
        \frame{\includegraphics[trim=3 3 3 3,clip,width=20mm]{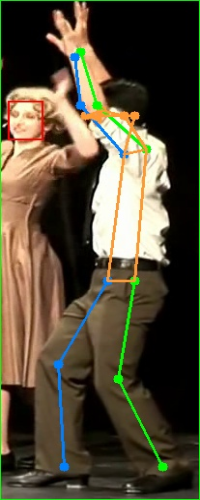}}
        \\
        \includegraphics[width=0.07\linewidth]{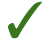} &
        \includegraphics[width=0.07\linewidth]{images/v-sign.pdf} &
        \includegraphics[width=0.07\linewidth]{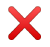} 
        \\
    \end{tabular}

    \vspace{-1em}

    \caption{\textbf{Matching Face Detection (red bounding box) and Pose Estimation (colored skeleton)}
    \textit{Left:} Detected face matches the pose estimation. 
    \textit{Middle:} Irrelevant detected faces are ignored as pose estimation matches a single face.
    \textit{Right:} Detected face does not match the pose estimation.}

    \label{fig:face_and_pose_match}
    \vspace{-0.5cm}

\end{figure}

A person crop is a part of an image that aims to capture a single individual. However, in crowded scenes multiple people might appear in the background, making it difficult to determine which face belongs to the main person in the crop. Therefore, when predicting an identity by using face features, it is crucial to verify that the detected face indeed belongs to the targeted individual. To achieve this, we utilize a pose estimator \cite{mmpose2020} (which we limit to provide a single pose estimation) to confirm that the eyes and nose keypoints of the main person in the crop fall within the face bounding box. Therefore, for datasets curated from crowded scenes (like the proposed \textbf{\textit{42Street}} dataset), \Cref{item:labeled,item:unlabeled} should include an additional step of face-to-pose matching. Examples are shown in \cref{fig:face_and_pose_match}.

\subsection{Combining ReID and Face Modules}
\label{subsec:combined-reid-and-face}
Following is a detailed description of our method for the video-based setting, where every data sample is a person track. The prediction process for the image-based setting is treated as a special case of the video-based setting, in which a data sample, \ie. a single image, is a track of length 1. In our method we propose to use a face feature extraction module and a ReID module to compute face and ReID score vectors respectively. These score vectors represent the confidence of each module that the given track 
belongs to each of the possible identities. Our method combines these two score vectors into a final score vector which is used to predict the identity of the person in a given data sample. \\
\begin{figure*}
    \includegraphics[width=\linewidth]{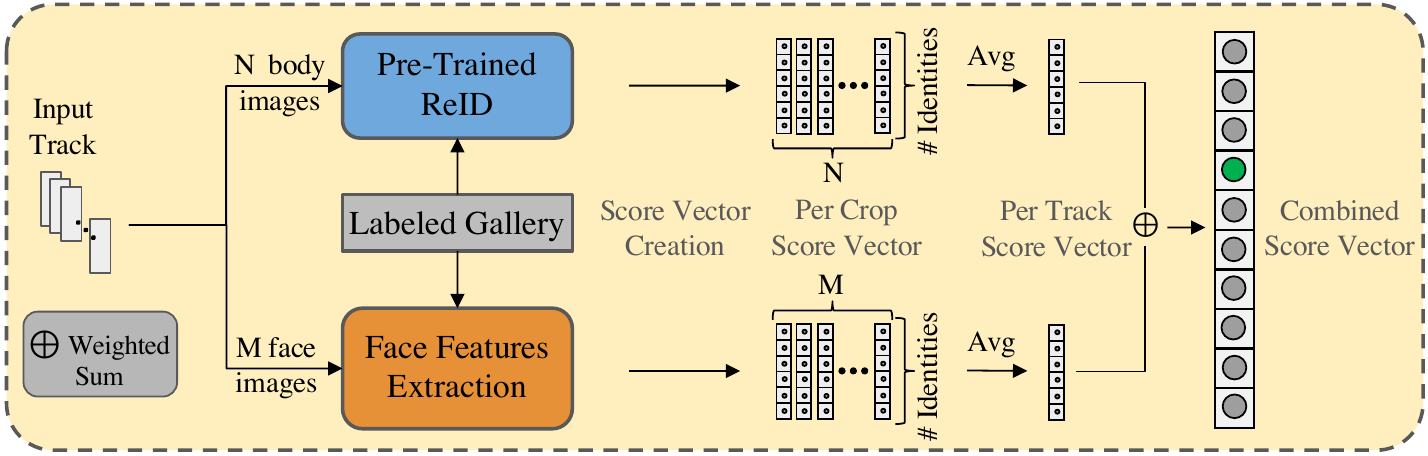}
    \caption{\textbf{Track Identity Prediction Overview.} 
    After receiving an input track, the ReID and face modules generate a track score vector that indicates the likelihood that the track belongs to each identity based on all track images. Subsequently, these score vectors are combined to form a conclusive score vector, and the track is identified by its highest score (indicated by a green circle).}
    \vspace{-0.3cm}
    \label{fig:track_to_score}
\end{figure*}

\textbf{Predicting an Identity of a Track} \label{our-model-video-based}
Given a gallery with a set of identities ${I}$, we first build an enriched gallery, \Genriched, as described in \cref{gallery-creation}, and a face gallery denoted \Gface, using face features extracted in the gallery enrichment process. 
Then, we use the ReID and face feature extraction modules on the track to create score vectors of size $|I|$, $v_{ReID} \in \mathbb{R}^{|I|}$ and $v_{face} \in \mathbb{R}^{|I|}$, respectively, which represent the confidence of each module that the given track belongs to each identity in ${I}$. This process is inspired by works such as CTL\cite{CTL} and MCTL\cite{MCTL} that use the identity of each sample during inference to calculate centroids. Finally, we combine $v_{ReID}$ and $v_{face}$ into a single score vector. The process is illustrated in \cref{fig:track_to_score} and detailed next.

\textbf{ReID and Face Score Vectors}  \label{face-reid-score-functions}
To compute $v_{ReID}$, for every crop $q$ in a track of size $N$, we compute the feature vector using the ReID module. Then, for each identity $i \in I$, we determine the confidence that image $q$ belongs to identity $i$ by finding the maximum cosine-similarity between the feature vector of $q$ and the feature vectors of all gallery samples of identity $i$ in \Genriched. The ReID score for identity $i$, \ie. $v_{ReID}[i]$, is the mean confidence for identity $i$ across all images in the track.

To compute $v_{face}$ we follow the same procedure as above while changing the input images and gallery. In this case, the input images for the procedure are $M$ detected face images from the original person track. Those are created by applying a face-detector on every image in the original track and comparing them to the gallery \Gface, while verifying that the pose matches the main person in the image (\cref{face-to-pose}).

\textbf{Combining Score Vectors} \label{combined-score}
The combined score vector is defined as:
\begin{equation} \label{score-vector-combination-equation}
v_{pred} = \alpha \cdot v_{ReID} + (1-\alpha) \cdot v_{face}
\end{equation}

With $\alpha$ as a hyper-parameter determining the weight of each module. In \cref{implementation-details}, we examine different $\alpha$ values. The final prediction for the entire track is given by taking the identity with the highest score in $v_{pred}$.

\section{The 42Street Dataset} \label{42 Street Dataset Creation}

Widely used clothes-changing ReID benchmarks (e.g. LTCC\cite{ltcc}, PRCC\cite{prcc}, LaST\cite{last}, VC-Clothes \cite{VC-Clothes}) do not capture crowded scenes that include multiple clothes changes per identity with various scale and illumination conditions. %
Moreover, CCVID\cite{ccvid}, a video-based clothes-changing ReID benchmark, was curated from a gait-recognition dataset (FVG \cite{gait-FVG}), in which the people are captured while walking towards the camera with clearly visible faces. On this dataset, a simple face-recognition model achieves superior results compared to ReID models, as shown in \cref{tab:table-ccvid}.
For these reasons, we publish a new video-based clothes-changing dataset --- the \textbf{\textit{42Street}} dataset. While the theater-play based dataset is attractive since it addresses
the challenges described above, there are not enough people-of-interest to create both training and evaluation sets. Hence, we publish it as an evaluation dataset only, which can be used as a new benchmark to test the generalization ability of ReID models. \\
The dataset is created using a public recording of the 42Street theatre play \cite{42street}.
The play is $\sim$1.5 hours long and we split it into 5 equally long parts of $\sim$20 minutes each, with various clothes changes between the different parts. From \textit{Part 1} we label samples from which a gallery is built. Parts \textit{2--3} and \textit{4--5} are used for validation and test, respectively. From these parts we randomly extract 5 validation videos and 10 test videos, each 17 seconds long.

\section{Experiments}

\subsection{Experiments on Existing Benchmarks} 
\begin{table}
    \centering
\begin{tabular*}{\linewidth}{@{\extracolsep{\fill}}{c}*{5}{c}}
\toprule
   \textbf{Name}  &\textbf{Type}  &\textbf{ID} &\textbf{Gallery} &\textbf{Query} & \textbf{Enrich}\\
\midrule
PRCC    & Img  &71    &3384   &7416 &2792  \\    
LTCC    & Img  &75    &7050    &493 &40\\
LaST    & Img  &5807   &125353  &10176 &4609\\
VC-Clothes  &Img  &256  &8591 &1020 &618 \\
CCVID   & Vid  &151   &1074   &834  &734\\
\bottomrule
\end{tabular*}
\caption {Statistics of the evaluation set of multiple clothes-changing ReID benchmarks. \textit{Enrich} shows the number of query samples used in the gallery enrichment process of each dataset. For the CCVID, the numbers indicate the number of sequences.}
\label{datasets-statistics}
\vspace{-0.5cm}
\end{table}

In this section, we show the potential improvement of several ReID models when applying GEFF, across all benchmarks detailed in \cref{datasets-statistics}. For each benchmark, we apply GEFF on the most recent ReID model for which we were able to reproduce similar results to the results reported in the original paper. Additionally, as our method relies on faces, we also apply a face-recognition model --- InsightFace\cite{deng2018arcface,deng2018menpo,Deng2020CVPR,deng2020subcenter}, to all the evaluated benchmarks as a face module baseline.

\subsubsection{Evaluation Protocol}\label{existing-benchmarks}  
Following most CC-ReID works, \textit{Top-1} accuracy and \textit{mAP} are used as evaluation metrics in three test scenarios:
\begin{itemize}[leftmargin=*]
    \item \textit{General:} All query and gallery samples are used to calculate accuracy.
    \item \textit{Clothes-Changing:} Uses only query samples that have matching gallery samples with \underline{different} clothes. Additionally, gallery samples with the same identity and the same clothes are discarded.
    \item \textit{Same-Clothes (SC):} Uses only query samples that have matching gallery samples with the \underline{same} clothes. Additionally, gallery samples with the same identity and different clothes are discarded.
\end{itemize}
Under all settings, gallery samples with the same identity and the same camera id are discarded.

\subsubsection{Results --- Existing Benchmarks} \label{benchmark-results}
In \cref{tab:table-method-improvment}, we apply GEFF to three ReID models on the PRCC and LTCC benchmarks showing an average improvement to the \textit{Top-1} and \textit{mAP} metrics, under most evaluated settings. 
Specifically, our method achieves an average improvement of 33.5\% and 6.7\%, under the clothes-changing setting, respectively.
\cref{tab:table-ltcc-prcc,tab:table-last,tab:table-vc-clothes} show that when applying GEFF on the SOTA models on the PRCC, LTCC, LAST, and VC-Clothes datasets, new SOTA results are achieved. On the CCVID benchmark, as the SOTA model DCR-ReID requires pre-processing of the dataset which is yet to be published, we apply our method on the second best model --- CAL. \cref{tab:table-ccvid} shows that applying GEFF achieves a significant improvement (outperforming even the DCR-ReID model). Notice that for this dataset, the baseline face-recognition model, InsightFace, achieves superior results under the general setting. We attribute this success to the fact that the CCVID dataset was curated from a gait recognition dataset, in which every image includes a clearly visible face. While the face-recognition model is successful on this dataset, the results on the other datasets, suggest that it is insufficient by itself.

\captionsetup{skip=0pt}
\begin{table*}
	\centering
	\begin{center}
	\begin{adjustbox}{width=1.0\textwidth}
    		\begin{tabular}{lcccccccccc}
    			\toprule
                    & &
    			&\multicolumn{4}{c}{PRCC}
    			&\multicolumn{4}{c}{LTCC} \\
    			\cmidrule(l{2pt}r{2pt}){4-7} \cmidrule(l{2pt}r{2pt}){8-11}
                    \textit{Method} & \textit{GE} & \textit{FF}\\[-0.3em] \vspace{0.1em} &  &
                    &\multicolumn{2}{c}{Same-Clothes} 
    			&\multicolumn{2}{c}{Clothes-Changing}
    			&\multicolumn{2}{c}{General} 
    			&\multicolumn{2}{c}{Clothes-Changing}\\
    			\cmidrule(l{2pt}r{2pt}){4-5} \cmidrule(l{2pt}r{2pt}){6-7} \cmidrule(l{2pt}r{2pt}){8-9} \cmidrule(l{2pt}r{2pt}){10-11}
    			& & &top-1 &mAP &top-1 &mAP &top-1 &mAP &top-1 &mAP\\
    			\midrule
                    & & &81.9 &73.0 &28.1 &23.4 &36.7 &11.1 &11.7 &4.8 \\
                    \it{CTL}\cite{CTL} &\checkmark & &93.8 (\textcolor{green}{+11.9}) &68.0 (-5.0) &78.5 (\textcolor{green}{+50.4})  &33.8 (\textcolor{green}{+10.4}) &38.9 (\textcolor{green}{+2.2}) &11.2 (+0.1) &16.3 (\textcolor{green}{+4.6}) &4.9 (+0.1) \\
                    &\checkmark &\checkmark &97.3 (\textcolor{green}{+15.4}) &76.6 (\textcolor{green}{+3.6}) &80.4 (\textcolor{green}{+52.3})  &42.3 (\textcolor{green}{+18.9}) &40.8 (\textcolor{green}{+4.1}) &12.0 (+0.9) &19.9 (\textcolor{green}{+8.2}) &5.5 (+0.7) \\
                    \midrule
    			     & &   &100 &99.8 &55.7 &56.3 &74.4 &41.2 &39.3 &19.0 \\
                    \it{CAL}\cite{CAL}    &\checkmark & &99.7 (-0.3) &99.5 (-0.3) &82.2 (\textcolor{green}{+26.5}) &59.3 (\textcolor{green}{+3.0}) &75.1 (+0.7) &41.2 (0) &45.4 (\textcolor{green}{+6.1}) &19.2 (+0.2) \\
                      &\checkmark &\checkmark &99.6 (-0.4) &99.4 (-0.4) &83.5 (\textcolor{green}{+27.8}) &64.0 (\textcolor{green}{+7.7}) &75.5 (\textcolor{green}{+1.1}) &41.8 (+0.6) &46.4 (\textcolor{green}{+7.1}) &20.2 (\textcolor{green}{+1.2}) \\
                    \midrule
    			& &  &100 &99.8 &58.2 &58.0 &75.9 &41.7 &40.8 &19.2 \\
    			\it{AIM} \cite{reid_aim}   &\checkmark &  &99.7 (-0.3) &99.4 (-0.4) &82.2 (\textcolor{green}{+24.0}) &60.4 (\textcolor{green}{+2.4}) &76.3 (+0.4) & 41.7 (0) & 45.2 (\textcolor{green}{+4.4}) &19.3 (+0.1)\\
                &\checkmark &\checkmark &99.8 (-0.2) &99.1 (-0.7) &82.5 (\textcolor{green}{+24.3}) &64.7 (\textcolor{green}{+6.7}) &76.3 (+0.4) &42.3 (+0.6) &45.7 (\textcolor{green}{+4.9}) &20.3 (\textcolor{green}{+1.1}) \\
                \midrule
                &\checkmark & &\textcolor{green}{+3.7} &-1.9  &\textcolor{green}{+33.6} &\textcolor{green}{+5.2} &\textcolor{green}{+1.1} &0 &\textcolor{green}{+3.2} &+0.1 \\
               \it{Avg.} &\checkmark &\checkmark &\textcolor{green}{+4.9} &+0.8  &\textcolor{green}{+34.6} &\textcolor{green}{+11.1} &\textcolor{green}{+1.8} &+0.7 &\textcolor{green}{+6.7} &\textcolor{green}{+1.0} \\

    			\bottomrule
    	    \end{tabular}
    \end{adjustbox}
    \end{center}
	\caption{\textbf{Applying GEFF on 3 ReID models over the PRCC and LTCC benchmarks.} In green are improvements of at least \textcolor{green}{+1.0\%}. The first row of every evaluated model is a result reproduced by us, done in order to create a fair comparison between a ReID model that we trained and the improvement achieved by applying GEFF on it.
    The second and third rows show the improvement achieved by the gallery enrichment (GE) and by applying GEFF (Enriched Gallery + Face Module), respectively. The last rows (Avg.) show the average improvement of applying an Enriched Gallery and GEFF across all models.}
	\label{tab:table-method-improvment}
\end{table*}

\begin{table*}[!b]
    \centering
    \setlength\tabcolsep{5pt}
    \setlength\extrarowheight{2pt} 
    \begin{tabular*}{\textwidth}{@{\extracolsep{\fill}}ll*{9}{c}}
        \toprule
        &
        \multirow{3}*{Method}
        & 
        &\multicolumn{4}{c}{PRCC} 
        &\multicolumn{4}{c}{LTCC}\\
        \cline{4-7} \cline{8-11}
        &
        &Year
        &\multicolumn{2}{c}{Same-Clothes} 
        &\multicolumn{2}{c}{Clothes-Changing}
        &\multicolumn{2}{c}{General} 
        &\multicolumn{2}{c}{Clothes-Changing}\\
        \cline{4-5} \cline{6-7} \cline{8-9} \cline{10-11}
        & & &top-1 &mAP &top-1 &mAP &top-1 &mAP &top-1 &mAP \\
        \midrule
        \multirow{5}*{\rotatebox{90}{Non-CC-Models}}
        & \textit{HACNN}\cite{HACNN}          &2018 &82.5 &- &21.8 &- &60.2 &26.7 &21.6 &9.3\\
        & \textit{PCB}\cite{PCB}         &2018 &99.8 &97.0 &41.8 &38.7 &65.1 &30.6 &23.5 &10.0\\
        & \textit{ISP}\cite{ISP}         &2020 &92.8 &- &36.6 &- &66.3 &29.6 &27.8 &11.9\\
        & \textit{InsightFace}  \cite{deng2018arcface,deng2018menpo,Deng2020CVPR,deng2020subcenter}      &2020 &95.6 &70.0 &78.0 &54.7 &27.4 &9.2 &24.5 &8.3\\
        & \textit{CTL} \cite{CTL}         &2021 &81.9 &73.0 &28.1 &23.4 &36.7 &11.1 &11.7 &4.8\\

        \midrule
        \multirow{8}*{\rotatebox{90}{CC-Models}}
                & \textit{FSAM}\cite{FSAM}          &2021 &98.8 &- &54.5 &- &73.2 &35.4 &38.5 &16.2\\
                & \textit{GI-ReID}\cite{GI-ReID}          &2022 &86.0 &- &33.3 &- &63.2 &29.4 &23.7 &10.4\\
                & \textit{UCAD}\cite{UCAD}          &2022 &96.5 &- &45.3 &- &74.6 &34.8 &32.5 &15.1\\
                & \textit{CAL} \cite{CAL}         &2022 &\bf100 & 99.8 &55.2 &55.8  &74.2 &40.8 &40.1 &18.0\\
                & \textit{ACID}\cite{ACID}          &2023 &99.1 &99.0 &55.4 &\textbf{66.1} &65.1 &30.6 &29.1 &14.5\\
                & \textit{CCFA}\cite{CCFA}          &2023 &99.6 &98.7 &61.2 &58.4 &75.8 &\textbf{42.5} &45.3 &\bf22.1\\
                & \textit{DCR-ReID}\cite{DCR-ReID}          &2023 &100 &99.7 &57.2 &57.4 &76.1 &42.3 &41.1 &20.4\\
                & \textit{AIM} \cite{reid_aim}          &2023 &\bf100 &\bf99.9 &57.9 &58.3 &\textbf{76.3} &41.1 &40.6 &19.1\\
                
                \bottomrule			
                & \textit{\textbf{AIM + GEFF}}        &2023 &99.8 &99.1 &\textbf{82.5} &64.7  &\textbf{76.3} &42.3 &\textbf{45.7}  &20.3 \\
                
        \bottomrule
    \end{tabular*}
    \caption{\textbf{Results on the LTCC and PRCC benchmarks.} \textit{CC-Models} (\textit{Non-CC-Models}) are ReID models that were (not) designed specifically for the clothes-changing challenge. \textit{AIM + GEFF} is the AIM model combined with a gallery enrichment and face module. Our method introduces an improvement over the Clothes-Changing setting and achieves comparable result on the Same-Clothes setting. }
    \label{tab:table-ltcc-prcc}
\end{table*}

\captionsetup{skip=2pt}
\begin{table}
	\centering
    \setlength\tabcolsep{5pt}
    \setlength\extrarowheight{2pt} 
	\begin{center}
            \begin{tabular*}{\linewidth}{@{\extracolsep{\fill}}{l}*{2}{c}}
			\toprule
			Method & {top-1} & {mAP} \\
			\midrule
			\textit{CTL} \cite{CTL}        &20.1 &3.2\\
   			\textit{InsightFace}\cite{deng2018arcface,deng2018menpo,Deng2020CVPR,deng2020subcenter} &57.8 &31.1 \\
            \textit{OSNet} \cite{OSNet}       &63.8 &20.9\\
            \textit{BoT}\cite{BoT}         &68.3 &25.3\\
            \textit{mAPLoss} \cite{last}    &69.9 &27.6\\
			\textit{CAL} \cite{CAL}        &73.7 &28.8 \\
            \midrule
			\textbf{\textit{CAL + GEFF}}        &\textbf{78.0} &\textbf{37.2} \\
			\bottomrule
	    \end{tabular*}
	\caption{\textbf{Results on LaST.} GEFF introduces a significant  improvement when applied to the CAL model.}
	\label{tab:table-last}
	\end{center}

\end{table}
\captionsetup{skip=2pt}
\begin{table}
	\centering
    \setlength\tabcolsep{1pt}
    \setlength\extrarowheight{2pt} 
	\begin{center}
		\begin{tabular*}{\linewidth}{@{\extracolsep{\fill}}{l}*{6}{c}}
			\toprule
			\multirow{2}*{Method}
			&\multicolumn{2}{c}{General} 
                &\multicolumn{2}{c}{SC}
                &\multicolumn{2}{c}{CC}\\
			\cmidrule(l{2pt}r{2pt}){2-3} \cmidrule(l{2pt}r{2pt}){4-5} \cmidrule(l{2pt}r{2pt}){6-7}
			&top-1 &mAP &top-1 &mAP &top-1 &mAP\\
			\midrule
                \textit{InsightFace}\cite{deng2018arcface,deng2018menpo,Deng2020CVPR,deng2020subcenter}     &83.8 &61.8 &92.7 &89.2 &63.1 &34.4\\
                \textit{PCB} \cite{PCB} &87.7 &74.6 &94.7 &94.3 &62.0 &62.2\\
                \textit{MDLA} \cite{MDLA} &88.9 &76.8 &94.3 &93.9 &59.2 &60.8\\
                \textit{3APF} \cite{VC-Clothes} &90.2 &82.1 &- &- &- &-  \\
                \textit{Part-align}\cite{part-aligned} &90.5 &79.7 &93.9 &93.4 &69.4 &67.3\\
                \textit{FSAM} \cite{FSAM} &- &- &94.7 &94.8 &78.6 &78.9\\
                \textit{3DSL} \cite{3DSL} &- &- &- &- &79.9 &81.2 \\
                
			\textit{CAL} \cite{CAL}        &92.9 &87.2 &95.1 &95.3 &81.4 &81.7\\ 
           \midrule
			\textbf{\textit{CAL+GEFF}}        &\textbf{94.9} &\textbf{88.9} &\bf96.5 &\textbf{96.3} &\textbf{86.7} &\textbf{84.4}\\
			\bottomrule
	    \end{tabular*}
	\caption{\textbf{Results on the VC-Clothes benchmark.} GEFF introduces a significant improvement under all settings when applied to the CAL model.}
	\label{tab:table-vc-clothes}
	\end{center}

\end{table}

\captionsetup{skip=2pt}
\begin{table}
	\centering
    \setlength\tabcolsep{5pt}
    \setlength\extrarowheight{2pt} 
	\begin{center}
		\begin{tabular*}{\linewidth}{@{\extracolsep{\fill}}{l}*{2}{c}}
			\toprule
			\multirow{2}*{Method}
			&\multicolumn{1}{c}{General} 
                &\multicolumn{1}{c}{Clothes-Changing}\\
			\cmidrule(l{2pt}r{2pt}){2-2} \cmidrule(l{2pt}r{2pt}){3-3}
			&top-1 &top-1\\
			\midrule
			\textit{CTL} \cite{CTL}       &71.8 &69.3 \\
            \textit{I3D} \cite{I3D}        &79.7 &78.5\\
            \textit{Non-Local} \cite{wang2018nonlocal}  &80.7 &79.3 \\
            \textit{AP3D} \cite{AP3D}        &80.9  &80.1 \\
            \textit{TCLNet} \cite{TCLNet}      &81.4  &80.7  \\
			\textit{CAL}\cite{CAL}         &82.9  &81.9  \\
           \textit{DCR-ReID} \cite{DCR-ReID}  &84.7  &83.6 \\
            \textit{InsightFace}\cite{deng2018arcface,deng2018menpo,Deng2020CVPR,deng2020subcenter}     &\bf95.3 &73.5 \\
           \midrule
			\textbf{\textit{CAL + GEFF}}        &89.2 &\bf90.5\\
			\bottomrule
	    \end{tabular*}
	\caption{\textbf{Results on the CCVID benchmark.} GEFF introduces a significant improvement when applied to the CAL model. The face model baseline (InsightFace) achieves a superior result as most tracks in this dataset include a clearly visible face image. As we explain in \cref{calculating-mAP}, mAP is not computed for video-based benchmarks when using our method.}
	\label{tab:table-ccvid}
	\end{center}
    \vspace{-0.5cm}
\end{table}

\textbf{Cross-Dataset Results} \label{cross-dataset-experiment}
In this experiment, we analyze the generalization ability of different models when training on one dataset but testing on another. \cref{cross-datasets-table} shows that applying GEFF to three ReID models increases their generalization abilities. While training the ReID models on the PRCC dataset and evaluating them on the LTCC dataset, applying GEFF achieves an average \textit{Top-1} improvement of 4.7\% and 6.4\% on the General and Clothes-Changing settings, respectively. While training the ReID models on the LTCC dataset and evaluating them on the PRCC dataset, applying GEFF achieves an average \textit{Top-1} improvement of 5.2\% and 44.5\% on the Same-Clothes and Clothes-Changing settings, respectively. Note that in both experiments, CTL was trained on the DukeMTMC\cite{duke-mtmc} dataset.
\captionsetup{skip=0pt}
\begin{table*}
	\centering
    \setlength\tabcolsep{3pt}
    \setlength\extrarowheight{2pt} 
	\begin{center}
	\begin{adjustbox}{width=1.0\textwidth}
    		\begin{tabular}{lcccccccc}
    			\toprule
    			\multirow{3}*{Method}
    			&\multicolumn{4}{c}{PRCC $\rightarrow$ LTCC}
    			&\multicolumn{4}{c}{LTCC $\rightarrow$ PRCC} \\
    			\cmidrule(l{2pt}r{2pt}){2-5} \cmidrule(l{2pt}r{2pt}){6-9}
    			&\multicolumn{2}{c}{General} 
    			&\multicolumn{2}{c}{Clothes-Changing}
    			&\multicolumn{2}{c}{Same-Clothes} 
    			&\multicolumn{2}{c}{Clothes-Changing}\\
    			\cmidrule(l{2pt}r{2pt}){2-3} \cmidrule(l{2pt}r{2pt}){4-5} \cmidrule(l{2pt}r{2pt}){6-7} \cmidrule(l{2pt}r{2pt}){8-9}
    			&top-1 &mAP &top-1 &mAP &top-1 &mAP &top-1 &mAP\\
                \midrule
                \it{CTL} &36.7 &11.1 &11.7 &4.8 &81.9 &73.0 &28.1 &23.4 \\
    			\it{\textbf{CTL + GEFF}}  &40.8 (\textcolor{green}{+4.1}) &12.0 (+0.9) &19.9 (\textcolor{green}{+8.2}) &5.5 (+0.7) &97.3 (\textcolor{green}{+15.4}) &76.6 (\textcolor{green}{+3.6}) &80.4 (\textcolor{green}{+52.3}) &42.3 \textcolor{green}{(+18.9)}  \\
       
    			\midrule
    			\it{CAL}            &21.9 &6.1 &7.4 &3.3 &99.6 &96.4 &37.7 &35.4 \\
                \it{\textbf{CAL + GEFF}}     &25.2 (\textcolor{green}{+3.3}) &6.1 (0) &11.5 (\textcolor{green}{+4.1}) &3.3 (0) 
                &99.7 (+0.1) &94.8 (-1.5) &80.7 (\textcolor{green}{+43.0}) &48.9 (\textcolor{green}{+13.5})\\
                \midrule
    			\it{AIM}            &22.1 &6.3 &6.1 &3.0 &99.6 &95.8 &40.7 &38.4 \\
    			\it{\textbf{AIM + GEFF}}     &29.2 (\textcolor{green}{+7.1}) &7.3 (\textcolor{green}{+1.0}) &13.0 (\textcolor{green}{+6.9}) &3.8 (+0.8) &99.7 (+0.1) &94.6 (-1.2) &81.0 (\textcolor{green}{+40.3}) &50.5 (\textcolor{green}{+12.1})\\
                    \midrule
                    \it{Avg.}            &\textcolor{green}{+4.7} &0.6 &\textcolor{green}{+6.4} &+0.5 &\textcolor{green}{+5.2} &0.3 &\textcolor{green}{+45.2} &\textcolor{green}{+18.8} \\
    			\bottomrule
    	    \end{tabular}
    \end{adjustbox}
    \end{center}
	\caption{\textbf{Cross-Dataset Generalization.} 
 $X \rightarrow Y$ means that the model was trained on dataset X and evaluated on dataset Y, with the exception of CTL that was trained on DukeMTMC. These results suggest that the generalization ability of ReID models increases significantly when applying the proposed GEFF method. The last row (Avg.) shows the average improvement of applying GEFF.}
	\label{cross-datasets-table}
	
	\vspace{-.3cm}
\end{table*}

\subsection{Experiments on The 42Street Dataset}\label{eval-42street}

\subsubsection{Evaluation Protocol}
Given the gallery created from \textit{part 1} of the play, we apply our method to the CTL ReID model and measure its performance on the evaluation videos of the \textbf{\textit{42Street}} dataset.
In our evaluation protocol, all models are evaluated without any training, as training data is not provided in this dataset.

\paragraph{Evaluation Metrics}

Similar to most ReID works, we measure the performance of a model using the top-1 metric. Since the query tracks are of different lengths, we measure the top-1 accuracy of both \textit{Per-Image} and \textit{Per-Track} accuracy.
\begin{itemize}[leftmargin=*]
    \item \textit{Per-Image Accuracy}:
the number of correctly identified person \underline{crops} in a video, divided by the total number of person crops in the video, across all evaluation videos.
\item \textit{Per-Track Accuracy}: the number of correctly classified \underline{tracks}, \ie. whether the model's single prediction on the entire track is correct, divided by the total number of tracks in the video, across all evaluation videos.
\end{itemize}
 
We observe that the image-based models we assess generate individual predictions for each image and do not offer a single prediction for an entire track. To overcome this shortcoming, we establish a single prediction based on the majority vote for the entire track. 
Additionally, since the primary focus of this study is not on enhancing tracking capabilities, we exclude tracks with less than 10 frames from our evaluation calculations, as they are more likely to be tracking errors. We note that the evaluation protocol of the \textbf{\textit{42Street}} dataset concerns only the detected person crops. The person detector of ByteTrack \cite{byteTracker} which we used, achieved an IDF1 score of 80.5 on MOT16\cite{MOT16}.

\subsubsection{Results --- 42Street Dataset}
\begin{figure}
    \centering
    \setlength{\tabcolsep}{5pt}
    \begin{tabular}{cc}
        \includegraphics[clip,width=75mm]{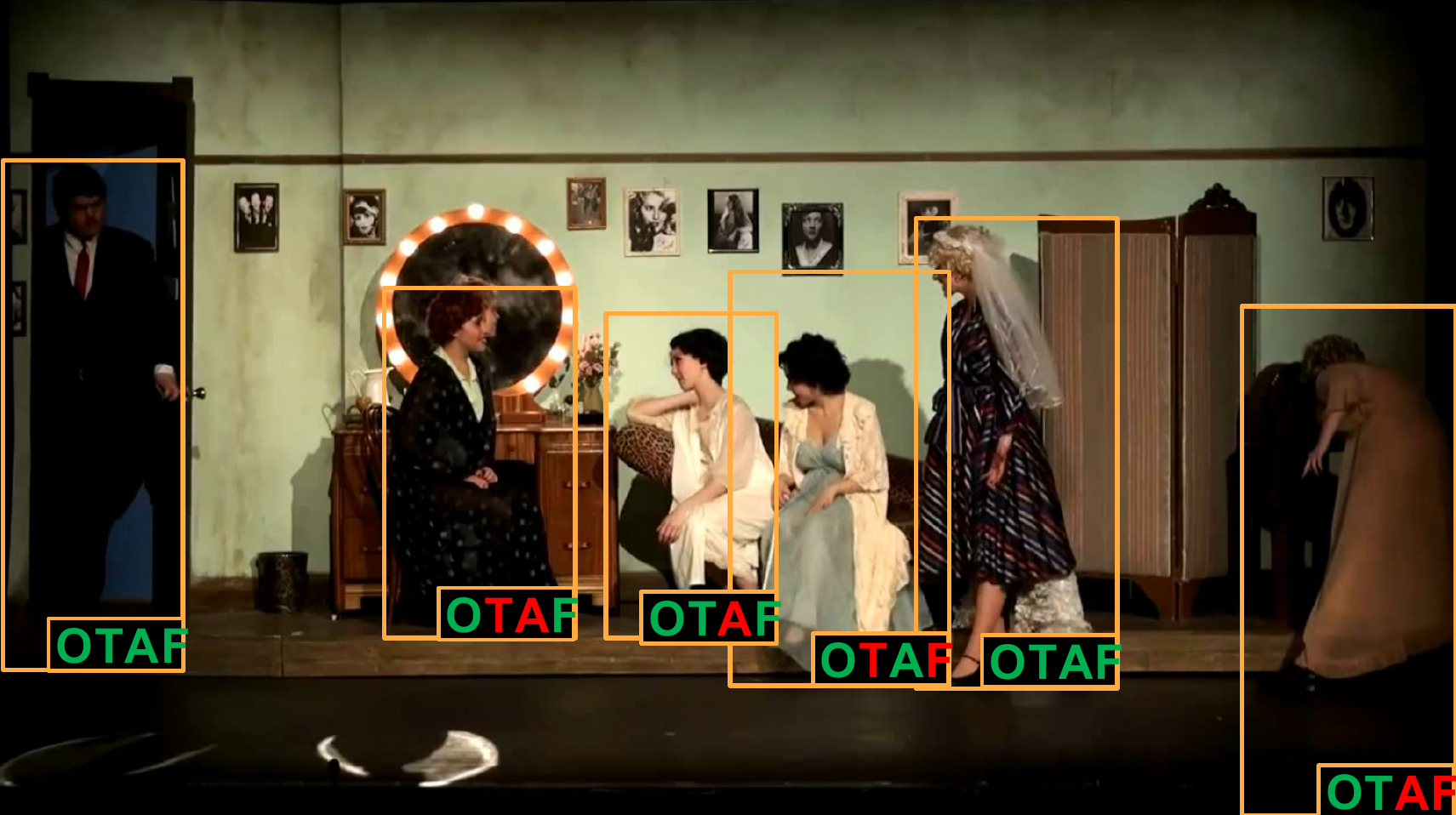} &
    \end{tabular}
    \caption{\textbf{Performance Visualization on the 42Street Dataset.}
    For each bounding box, the abbreviation 'OTAF' refers to the different models --- O (Ours), T (CTL), A (CAL), F (InsightFace). The green and red colors correspond to a correct and incorrect prediction of each model on the bounding-box.
    } 

    \label{fig:42_street_example}

\end{figure}

\cref{tab:42street-overview-results_test} summarizes the results on the \textbf{\textit{42Street}} dataset.\\ 
Similarly to the results shown in \cref{cross-dataset-experiment}, since the evaluated models have a limited generalization ability, they perform poorly on this dataset as they are not being trained on it. However, when applying GEFF to the CTL and CAL ReID models, whilst not requiring any further training on the dataset, strong results are achieved, significantly outperforming the baseline ReID models. 
\cref{fig:42_street_example} visualizes the performance of the various models on a single frame from a test video.
Interestingly, face-recognition by itself achieves mediocre results, as the face detector detected faces only in 74\% of the total person crops, some of which are not of sufficient quality to be recognized correctly. 

\section{Ablation Study} \label{ablation_study}
In this section, we evaluate the impact of each component of GEFF on the overall performance of a ReID model, both on the existing benchmarks and the \textbf{\textit{42Street}} dataset.
From \cref{tab:table-method-improvment,tab:42street-ablation} we conclude that
the \textit{Gallery Enrichment Process} introduces a significant improvement compared to using only the original gallery. In \cref{enrichment-percentage}, we further discuss the influence of additional raw data in the gallery enrichment process on the performance of the ReID model. Moreover, combining ReID and face modules using score vectors (even without gallery enrichment), significantly improves the results of ReID models.
Finally, although the face module achieves solid results on some benchmarks (showing the significance of face features for ReID tasks), it is an insufficient model by itself as biometric information such as faces is often unavailable in ReID problems.

\captionsetup{skip=2pt}

\begin{table}
\centering

\setlength\tabcolsep{5pt}
\begin{tabular*}{\linewidth}{@{\extracolsep{\fill}}{l}*{2}{c}}
\toprule
\multirow{2}*{Method} & \multicolumn{2}{c}{top-1} \\
\cmidrule(l{2pt}r{2pt}){2-3} & Image & Track  \\ 
\midrule
\it{InsightFace}          & 54.8 & 62.2   \\ 
\midrule
\it{CAL}                  & 22.2 & 21.5  \\
\textbf{CAL + GEFF}       & 74.1 \textcolor{green}{(+51.9)} & 66.7 \textcolor{green}{(+42.2)}  \\
\midrule
\it{CTL}                  & 31.1 & 26.7  \\
\textbf{CTL + GEFF}  & 91.9 \textcolor{green}{(+59.1)} & 81.8 \textcolor{green}{(+51.1)} \\ 

\bottomrule

\end{tabular*}
 \caption {\textbf{Results on the 42Street dataset.} All models are pre-trained on other datasets and are not fine-tuned on this dataset. 
 \textit{CTL, CAL} and \textit{InsightFace} are image-based models, for which we apply a majority vote in order to calculate per-track accuracy.
 }
    \label{tab:42street-overview-results_test}

\end{table}

\begin{table}[t]
\centering
\begin{tabular*}{\linewidth}{@{\extracolsep{\fill}}{l}*{2}{c}}
\toprule
\multirow{2}*{Method} & \multicolumn{2}{c}{top-1} \\ 
\cmidrule(l{2pt}r{2pt}){2-3}
\multicolumn{1}{c}{}          & \multicolumn{1}{c}{Image } & \multicolumn{1}{c}{Track }  \\ 
\midrule
\it{ReID}   & 31.1 & 26.7 \\
\it{+ Enriched}   & 80.1 & 71.1    \\
\it{Face}   & 54.8 & 62.2  \\
\it{ReID + Face } & 81.3 & 66.7   \\
\it{+ Enriched (\textbf{GEFF})} & \textbf{90.5} & \textbf{77.8} \\ 
\bottomrule
\end{tabular*}
 \caption {\textbf{Ablation study of GEFF on the 42Street dataset.} 
 The used ReID module is CTL. 
 }
    \label{tab:42street-ablation}
    \vspace{-.3cm}
\end{table}

\section{Ethical Considerations}
New person ReID and tracking datasets raise privacy concerns as individuals may appear in them without consent. In this work, we use publicly available videos from the \textbf{\textit{42 Street}} theatre play and only utilize face features for image retrieval and distance measurement, without identity matching. Our dataset is intended for academic use only. Moreover, we condemn the usage of ReID methods with nefarious intent and publish this work to progress academic research in this field.

\section{Limitations \label{limitations}}
    In order to enrich the gallery with samples of an unseen clothes-set (of a single identity), the gallery enrichment process relies on the assumption that at least one sample with these clothes includes a clearly visible face. For datasets where this assumption does not hold on multiple clothes-sets (e.g. LTCC), applying GEFF would only introduce a slight improvement, as only a limited amount of query samples will be added to the original gallery. 
    That said, we believe our assumption holds for many real-world scenarios and as such can introduce a significant improvement when applied to ReID models, as we showed on multiple datasets.
    
    Additionally, to apply our work to real-world applications we use a tracking module to extract person tracks. Therefore, we inherit all the tracker's limitations, such as missed detection and mid-track identity switches. We use the tracking module without applying any changes to it, thus we do not deal with these potential tracking mistakes.
    
Finally, the proposed \textbf{\textit{42Street}} dataset does not include a training set with a separate identity set, as the number of identities in the data was limited. Hence, it should be used for evaluation purposes only. Moreover, the dataset does not conform to the standard CC-ReID dataset settings, as it does not provide clothes and camera ids labels. However, we note that the gallery and query samples are taken from different (non-overlapping) parts of the play, captured with dynamic camera settings (various scene cuts, angles, and scales), and multiple clothes sets per identity. Therefore, we find this dataset a valid CC-ReID benchmark and an important contribution to the field, especially since the number of publicly available video CC-ReID datasets is limited.

\section{Conclusion}
In this work, we show a simple yet effective approach to address the clothes-changing ReID challenge by creating an enriched gallery from the given query and gallery samples. By applying GEFF on existing ReID models, new SOTA results are achieved, both on the existing clothes-changing ReID benchmarks and on the real-world clothes-changing dataset we publish, \textit{\textbf{42Street}}. Furthermore, we showed that by using GEFF, the generalization ability of existing ReID models increases, without requiring any further training.

\section{Acknowledgements}
This project was partially supported by ISF grant No. 1574/21
\appendix

\section{Using Raw-Data For Gallery Enrichment}\label{enrichment-percentage}
\begin{figure*}
    \centering
    \includegraphics[clip,width=80mm]{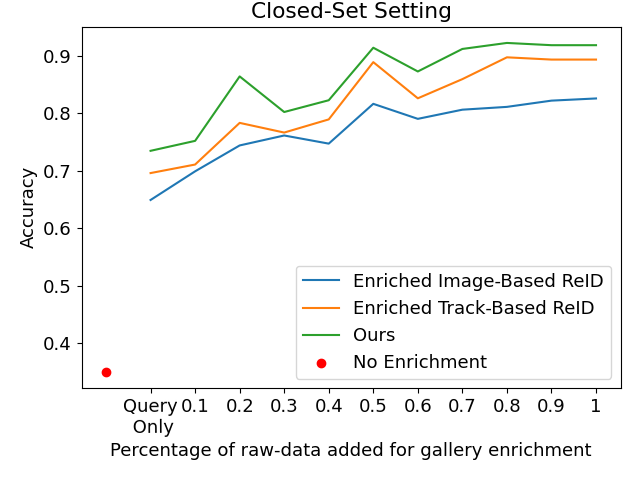}
    \includegraphics[clip,width=80mm]{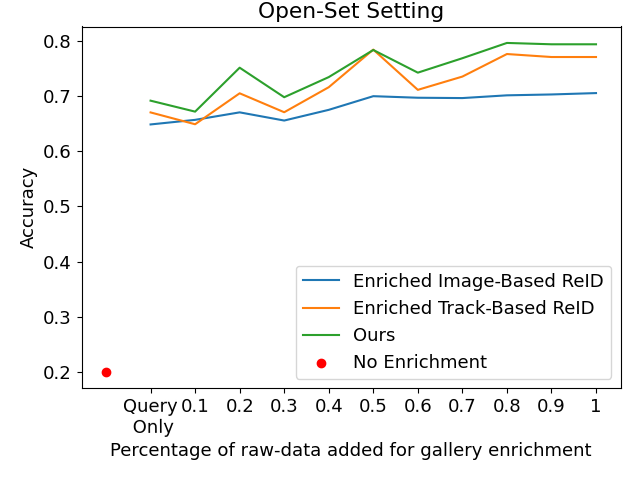}
    
    \caption{\textbf{Accuracy vs. Additional Raw Data for Gallery Enrichment in \textit{42Street}} show the relationship between accuracy and the percentage of additional raw data utilized for gallery enrichment in the \textit{42Street} dataset. The impact of ReID module on closed-set (left) and open-set (right) settings is demonstrated by enriching the gallery with extra raw data. The X-axis indicates the percentage of raw-data used for gallery enrichment. The red point represents image-based ReID module results without any enrichment. The \textit{Query Only} denotes using data solely from evaluation videos for gallery enrichment. Each step shows the percentage of extra data from the raw-video.
    The findings suggest that enriching the gallery with the \textit{Query Only} data significantly enhances the ReID module compared to no enrichment. Additionally, using more raw-data improves the overall accuracy of all compared modules.}
    \vspace{-0.3cm}
    \label{fig:query-enrichment-growth}
\end{figure*}

As explained in the \textbf{\textit{42Street}} dataset creation, the test data of this dataset consists of two parts from the play, each around 20 minutes long. Out of these parts,  we extract short videos, 17-seconds each, for evaluation. In this section, in addition to enriching the gallery with the query data (from the evaluation videos), we analyze the impact of using more raw-data (from the test part) for the gallery enrichment process. We start by using only the evaluation videos for gallery enrichment, and gradually add more randomly sampled raw data from the test parts. \cref{fig:query-enrichment-growth} shows the accuracy of the compared models as a function of the amount of raw-data used for the gallery enrichment process. Notice that the initial gallery enrichment with query data only, already introduces a significant improvement compared to not using enrichment at all, and that the accuracy increases as we add more raw-data. This is true for the Image-based ReID model, Track-based ReID model and our method, on both the closed and open set settings, with the best results achieved by our method. Even though raw-data is not available in the standard ReID task, we argue that such data is common in real-world scenarios like the application we presented, and show how it can be used to boost the performance of our method.

\section{Construction of the \textbf{\textit{42Street}} Dataset} \label{database}
\subsection{Database Structure}
We save each labeled crop from the \textit{\textbf{42Street}} dataset in a designated database, which includes spatial and temporal information for each crop. Every entry in the database includes the following information: 
\begin{itemize}
    \setlength\itemsep{-0.2em}
    \item \textit{label}.
    \item \textit{im\_name}: unique entry.
    \item \textit{frame\_num}: frame number of crop in video.
    \item \textit{x1,y1,x2,y2}: top-left and bottom-right coordinates of the crop's bounding box.
    \item \textit{conf}: person detector's confidence that a person exists in the crop.
    \item \textit{vid\_name}: the name of the video.
    \item \textit{track\_id}: the track number given by the tracker.
    \item \textit{crop\_id}: crop number within the track. \item \textit{invalid}: boolean value set to true for crops that do not represent a clear person.
\end{itemize} 

\subsection{Annotating Evaluation Videos  \label{42street-evaluation-videos}}

To annotate an evaluation video with ground truth labels, a tracker, ByteTrack \cite{byteTracker}, is applied to automatically extract tracks of detected people, followed by a manual annotation of every track. These ground-truth labels, as well as person bounding boxes, track ids, video names, and more are saved in a designated database published alongside the dataset.

\section{Threshold Details}\label{threshold-details}
In \cref{tab:thresholds-benchmarks} we detail the detection and similarity thresholds used by our method for the existing benchmarks. These  thresholds were defined based on the training data of each benchmark.
Moreover, for the \textbf{\textit{42Street}} dataset, we used the evaluation data to extract the following thresholds:
\begin{itemize}
    \item \textit{Detection Threshold}: For the gallery enrichment process we used a threshold of 0.8 and during inference, we used 0.7. The reason we use a higher threshold for the gallery enrichment process is that we are interested in creating a highly accurate gallery. On the other hand, during inference, we want to use more images with faces, and we are more tolerant of mistakes.
    \item \textit{Similarity Threshold}: During the gallery enrichment process we use a threshold of 0.4 for the cosine similarity between an unlabeled sample and the labeled gallery. That is, if the maximal cosine similarity between the feature vector of the unlabeled sample and the labeled gallery is below this threshold, this sample will not be added to the enriched gallery.
    \item \textit{Rank Difference}: In addition to detecting only clearly visible faces with high similarity to the labeled gallery, we want to use only samples for which the face model was confident about their identity. We measure this confidence as the difference between the similarity score of the top-1 and top-2 predictions of the model. In the \textbf{\textit{42Street}} dataset, if the difference is below 0.1, we do not add the sample to the enriched gallery.
\end{itemize}
Under the open-set setting, to label people who are not in the people-of-interest set as ``Unknown", we set another threshold. This threshold asserts that a person will be labeled as ``Unknown" if the similarity to all people-of-interest is below 0.3. This means that whilst having high detection confidence, the model is not confident of the identity.
\begin{table*} 
	\centering
    \setlength\tabcolsep{5pt}
    \setlength\extrarowheight{2pt} 
	\begin{center}
		\begin{tabular*}{\textwidth}{@{\extracolsep{\fill}}{c}*{8}{c}}
			\toprule
			\multirow{3}*{$\alpha$}
			&\multicolumn{4}{c}{PRCC} 
			&\multicolumn{4}{c}{LTCC}\\
			\cline{2-5} \cline{6-9}

			&\multicolumn{2}{c}{Same-Clothes} 
			&\multicolumn{2}{c}{Clothes-Changing}
			&\multicolumn{2}{c}{General} 
			&\multicolumn{2}{c}{Clothes-Changing}\\
			\cline{2-3} \cline{4-5} \cline{6-7} \cline{8-9}
			&top-1 &mAP &top-1 &mAP &top-1 &mAP &top-1 &mAP \\
			\midrule
                
                \textit{0} &93.7 &64.5 &71.3 &47.7 &15.6 &6.5 &13.3 &6.3 \\
                \textit{0.25} &99.6 &82.8 &84.3 &62.5 &76.3 &37.0 &46.4 &18.8 \\
                \textit{0.5} &99.8 &95.7 &83.4 &65.2 &76.3 &40.6 &45.9 &19.8 \\
                \textit{0.75} &99.8 &99.1 &82.5 &64.7 &76.3 &42.3 &45.7 &20.3 \\
                \textit{1} &99.7 &99.4 &82.2 &60.4 &76.3 &41.7 &45.2 &19.3 \\

			\bottomrule
	    \end{tabular*}
	\caption{\textbf{Ablation study on the impact of different $\alpha$ values}. Alpha values of 0 and 1 are equivalent to using only the face and ReID modules, respectively. We conclude that an $\alpha$ of 0.75 presents a good balance between the weight given to the ReID and Face modules.}
	\label{tab:alpha_ablation}
	\end{center}
    \vspace{-0.5cm}

\end{table*}

\section{Implementation Details \label{implementation-details}}

\subsection{Setting Hyper-Parameters \label{hyper-parameters}} 
\begin{figure}
    \centering
    \includegraphics[width=\linewidth]{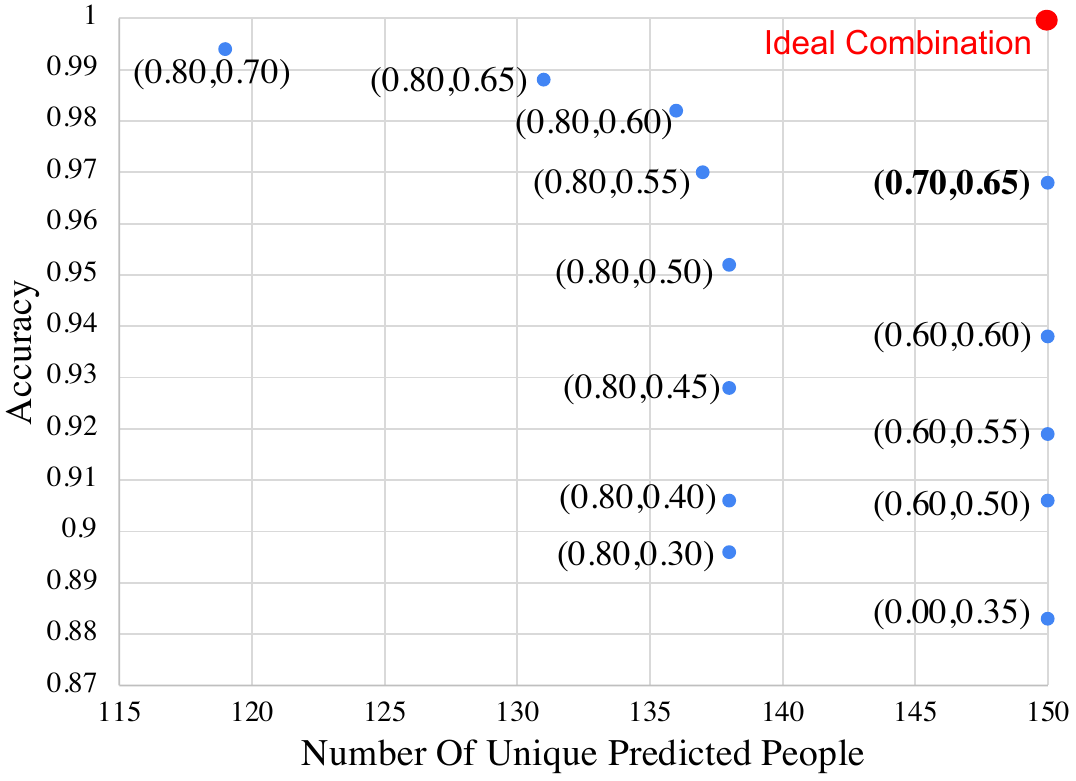}
    \caption{\textbf{Prediction Accuracy vs. Number of Unique People.} Example of the thresholds trade-off on the PRCC benchmark. Each data tuple represents the values of the detection threshold (left) and similarity threshold (right). We look for the combination that achieves the highest prediction accuracy while predicting the highest number of unique people (bold).}
    \label{fig:threshold-selection}
\end{figure}

A typical face-detection model produces a confidence score that a face exists in a given image. Additionally, cosine distances between the face feature vector of the input image and a face gallery are calculated. These distances can be used as the confidence that the given input image belongs to a certain identity. To achieve the best results, we set thresholds for both confidence scores.
Given a dataset, we apply our gallery enrichment method to the training set to find the best combination of these two thresholds. Ideally, we would like to find the optimal combination that achieves the highest prediction accuracy based on face features, whilst maintaining a high detection accuracy per person. Meaning, that we would like the model to predict at least one image of each identity in order to create an enriched gallery with examples of all identities. However, there exists a trade-off between the per-identity detection accuracy and the prediction accuracy: for a higher detection threshold, the number of unique predicted identities decreases. At the same time, since the detected faces are of better quality, the prediction accuracy increases. Vice versa, a lower detection threshold, leads to a larger number of predicted identities while achieving lower prediction accuracy.
\cref{fig:threshold-selection} demonstrates this trade-off for the PRCC benchmark.

\begin{table}
    \centering
    \begin{tabular}{lcc}
    \hline
    Threshold &
      \multicolumn{1}{l}{Detection} &
      \multicolumn{1}{l}{Similarity} \\
      \hline
      CCVID   &0.5  &0.75 \\
      LTCC    &0.85  &0.5  \\ 
      PRCC    &0.7  &0.65 \\
      LaST    &0.7  &0.45 \\
      \hline
    \end{tabular}

    \caption{\textbf{Threshold values for existing benchmarks.} The different detection and similarity thresholds used for the gallery enrichment process in each of the existing benchmarks.}
    \label{tab:thresholds-benchmarks}
    \vspace{-.3cm}
\end{table}

\begin{figure*}
    \centering
    \includegraphics[width=\linewidth]{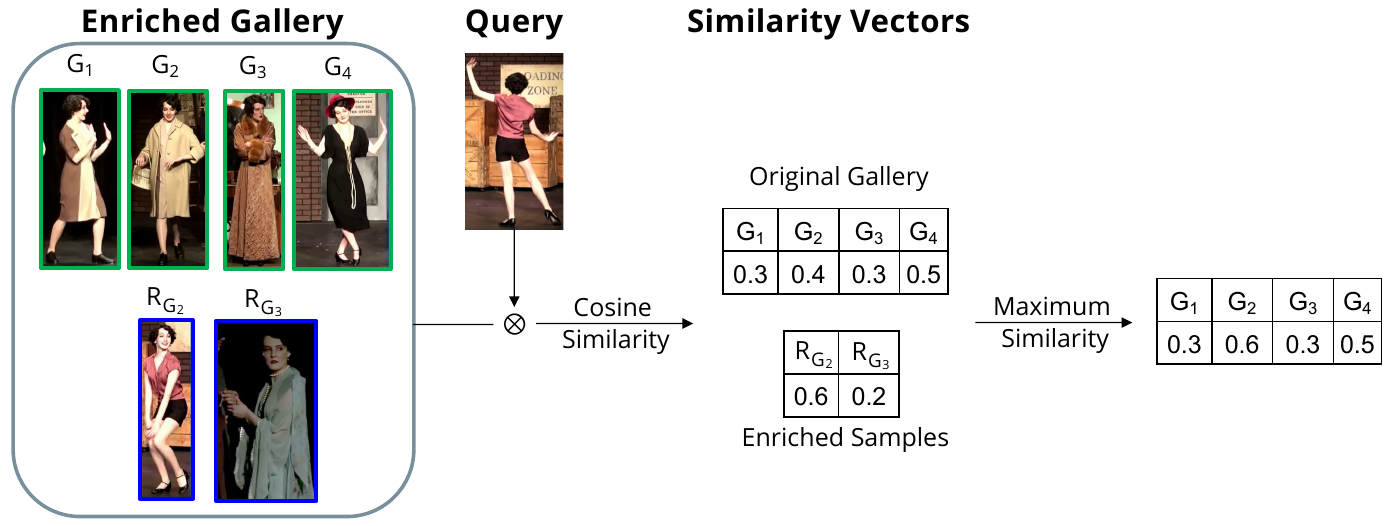}
    \caption{\textbf{Computing Similarities from an Enriched Gallery.} Given an enriched gallery and a query sample, we first compute similarity vectors between the original gallery samples (green frames) and the query (top similarity vector), and the enriched samples (blue frames) and the query (bottom similarity vector). Then, we combine the two similarity vectors into a final similarity vector in the size of the original gallery, by taking the maximum between each original sample similarity and the enriched samples that used it as a reference. For example, note that the similarity of gallery sample $G_{2}$ was replaced with the similarity of the enriched sample $R_{G_{2}}$ that used it as a reference in the gallery enrichment process.
    }
    \label{fig:enriched-gallery-similarity}
\end{figure*}

\subsection{Used Face and ReID Modules}
For our face module, we use \textit{InsightFace}  \cite{deng2018arcface,deng2018menpo,Deng2020CVPR,deng2020subcenter} for face detection, alignment, and feature extraction. For the ReID module, we examine both CTL\cite{CTL}, pre-trained on DukeMTMC \cite{duke-mtmc}, CAL\cite{CAL} and AIM\cite{reid_aim}.
The hyper-parameter $\alpha$ used for combining the face score vector and the ReID score vector is set to 0.75, giving more weight to the ReID module. An ablation model of the impact of $\alpha$ is presented in \cref{tab:alpha_ablation}. The detection and similarity thresholds used by our face module are determined according to the training set for the existing benchmarks, and the validation set for the \textbf{\textit{42Street}} dataset. In \cref{threshold-details}, we detail all the thresholds used for the different datasets.
\\
For \textbf{\textit{42Street}}, the entire raw videos of test-set parts are used as query input for the gallery enrichment process. To apply our method on the real-world application described in the paper, we use  ByteTrack \cite{byteTracker} as our tracking module as well as the MMCV \cite{mmcv} and MMTRACK \cite{mmtrack2020} frameworks. To deal with some of the tracking limitations, as the original play includes many scene cuts, i.e. abrupt changes of the camera angle or zoom, we apply a scene cut detection algorithm, and split tracks accordingly if necessary.

\subsection{Hardware}
First, we note, that our method does not require any training and uses only pre-trained ReID and face modules. However, since the evaluated ReID models do not release their trained checkpoints, we reproduced the results using the original code published by the authors. In this work, we used two GPUs: NVIDIA TITAN V  with 12GB memory and Quadro RTX 5000 with 16GB.

\section{Open-set Settings \label{open-set}}
\paragraph{Open-Set and Closed-Set} \label{open-vs-closed-sets}
A closed-set setting assumes that every identity in the query set has at least one corresponding sample in the gallery set\cite{Open-World-Survey}. In contrast, in an open-set setting, some identities in the query may not be present in the gallery. While closed-set ReID can be seen as an instance retrieval problem, open-set ReID is usually formulated as a person verification problem. In this formulation, the model is required to discriminate whether two person images belong to the same identity\cite{unsupervised-open-set, Fast-open}. Models that address the open-set scenario \cite{group-open,Group-Re-Identification, GAN-open-set} typically learn to discriminate between a given query and gallery images according to their similarity\cite{Open-World-Survey}. In this work, we use the gallery enrichment process to classify previously unseen people as ``Unknown". This allows us to recognize such query samples during inference and effectively enables models that originally addressed only the closed-set setting, to address the open-set setting. 

\paragraph{Addressing the Open-Set Challenge}\label{open-set-challenge} 
In the open-set setting, we approach the task of labeling an out-of-gallery query sample by utilizing the gallery enrichment process. During this process, we label a given query sample as ``Unknown'' and add it to the enriched gallery if it fulfills the following criteria:
\begin{enumerate}[leftmargin=*]
\item A face was detected.
\item The cosine similarity between the feature vector of the detected face and each of the labeled face gallery feature vectors is below a certain threshold.
\item The difference between the closest and second-closest predictions in the face score vector is below a certain threshold. 
\end{enumerate}

We evaluate this capability on the 42Street data set in \cref{tab:42street-open-set}. We limit this evaluation to the CTL model and our proposed dataset since, to the best of our knowledge, currently available CC-ReID benchmarks and ReID models operate under the closed-set setting.

\captionsetup{skip=10pt}

\begin{table}
\centering
\setlength{\tabcolsep}{2pt}
\begin{tabular}{lcccc}
\toprule
\multirow{2}*{Method} & \multicolumn{2}{c}{Closed-Set}                                         & \multicolumn{2}{c}{Open-Set} \\ 
\cmidrule(l{2pt}r{2pt}){2-3} \cmidrule(l{2pt}r{2pt}){4-5}
                   & Per Image & Per Track & Per Image & Per Track \\ 
\midrule
\it{CTL}             & 31.3 & 26.7                   & 20.5  & 15.5           \\
\\ \midrule
\it{\textbf{CTL + GEFF}}            & \textbf{91.9} & \textbf{81.8} & \textbf{80.5} & \textbf{65.2}   \\ 
\bottomrule

\end{tabular}
 \caption {\textbf{Results on the 42Street dataset under the closed/open set settings.} Applying our method to the pre-trained \textit{CTL} model, significantly improves the results of the model under both settings.
 }
    \label{tab:42street-open-set}

\end{table}

\section {Calculating mAP}\label{calculating-mAP}
In instance retrieval tasks, given a query sample, the goal of the model is to rank all gallery samples from the most similar to the least similar. The mAP metric measures the rank of all ``positive'' gallery samples for each query, i.e. the positions of all gallery samples with the same label as the query, compared to the positions of all other gallery samples. A model with a 100\% mAP score would rank the ``positive'' samples before all other gallery samples.
During the gallery enrichment process, we add samples to the gallery, hence resulting in a larger gallery than the original one. Therefore, in order to provide a fair comparison with previous works, during evaluation, we have to reduce the size of the gallery to its original size. 
For each query sample, a similarity vector is computed, holding the similarities between the query and all original gallery samples. Similarly, a similarity vector is computed between the query and all enriched samples. Finally, the similarity vectors are combined, resulting in a similarity vector of the same size as the original gallery. The combination is done by iterating over every original gallery sample and examining the group of all enriched samples that used it as a reference (i.e. this sample was the most similar gallery sample based on face features similarity) during the enrichment. Then, we set the similarity of the gallery sample in the similarity vector, as the maximum similarity between the query and the samples in this group including the original gallery sample. This process is illustrated in \cref{fig:enriched-gallery-similarity}.
We note that for the evaluated video benchmarks (CCVID, 42Street) our method utilizes score vectors to predict the identity of an entire track. The score vector holds a score per identity, and not ranking on the entire gallery. Therefore, mAP is not computed on these benchmarks.

{\small
\bibliographystyle{ieee_fullname}
\bibliography{egbib}
}

\end{document}